\documentclass[10pt,twocolumn,letterpaper]{article}

\usepackage{cvpr}
\usepackage{times}
\usepackage{epsfig}
\usepackage{graphicx}
\usepackage{amsmath}
\usepackage{amssymb}
\usepackage{algpseudocode}
\usepackage{multirow}

\cvprfinalcopy 

\def\cvprPaperID{418} 

\ifcvprfinal\fi
\begin{document}

\title{DeepPose: Human Pose Estimation via Deep Neural Networks}

\author{Alexander Toshev\quad\quad Christian Szegedy \\
Google\\
1600 Amphitheatre Pkwy \\
Mountain View, CA 94043 \\
{\tt\small {toshev,szegedy}@google.com}
}

\maketitle

\begin{abstract}
We propose a method for human pose estimation based on Deep Neural Networks (DNNs). The pose estimation is formulated as a DNN-based regression problem towards body joints. We present a cascade of such DNN regressors which results in high precision pose estimates. The approach has the advantage of reasoning about pose in a holistic fashion and has a simple but yet powerful formulation which capitalizes on recent advances in Deep Learning. We present a detailed empirical analysis with state-of-art or better performance on four academic benchmarks of diverse real-world images.
\end{abstract}

\section{Introduction}
\begin{figure}
{\centering
\includegraphics[width=0.19\textwidth,height=0.17\textwidth]{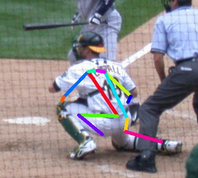}
\includegraphics[width=0.19\textwidth,height=0.17\textwidth]{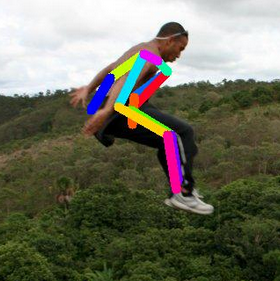}
\caption{\label{fig:intro_figure} Besides extreme variability in articulations, many of the joints are barely visible. We can guess the location of the right arm in the left image only because we see the rest of the pose and anticipate the motion or activity of the person. Similarly, the left body half of the person on the right is not visible at all. These are examples of the need for \textit{holistic reasoning}. We believe that DNNs can naturally provide such type of reasoning.}
}
\end{figure}

The problem of human pose estimation, defined as the problem of localization of human joints, has enjoyed substantial attention in the computer vision community. In Fig.~\ref{fig:intro_figure}, one can see some of the challenges of this problem -- strong articulations, small and barely visible joints, occlusions and the need to capture the context.

The main stream of work in this field has been motivated mainly by the first challenge, the need to search in the large space of all possible articulated poses. Part-based models lend themselves naturally to model articulations (\cite{nevatia1977description,fischler1973representation}) and in the recent years a variety of models with efficient inference have been proposed (\cite{felzenszwalb2005pictorial,ramanan2006learning}).

The above efficiency, however, is achieved at the cost of limited expressiveness -- the use of local detectors, which reason in many cases about a single part, and most importantly by modeling only a small subset of all interactions between body parts. These limitations, as exemplified in Fig.~\ref{fig:intro_figure}, have been recognized and methods reasoning about pose in a holistic manner have been proposed \cite{mori2002estimating,shakhnarovich2003fast} but with limited success in real-world problems.

In this work we ascribe to this holistic view of human pose estimation. We capitalize on recent developments of deep learning and propose a novel algorithm based on a Deep Neural Network (DNN). DNNs have shown outstanding performance on visual classification tasks \cite{krizhevsky2012imagenet} and more recently on object localization \cite{szegedy2013object,girshick2014rcnn}. However, the question of applying DNNs for precise localization of articulated objects has largely remained unanswered. In this paper we attempt to cast a light on this question and present a simple and yet powerful formulation of \textit{holistic human pose estimation} as a DNN.

We formulate the pose estimation as a joint regression problem and show how to successfully cast it in DNN settings. The location of each body joint is regressed to using as an input the full image and a 7-layered generic convolutional DNN. There are two advantages of this formulation. First, the DNN is capable of capturing the full context of each body joint -- each joint regressor uses the full image as a signal. Second, the approach is substantially simpler to formulate than methods based on graphical models -- no need to explicitly design feature representations and detectors for parts; no need to explicitly design a model topology and interactions between joints. Instead, we show that a generic convolutional DNN can be learned for this problem.

Further, we propose a cascade of DNN-based pose predictors. Such a cascade allows for increased precision of joint localization. Starting with an initial pose estimation, based on the full image, we learn DNN-based regressors which refines the joint predictions by using higher resolution sub-images. 

 We show state-of-art results or better than state-of-art on four widely used benchmarks against all reported results. We show that our approach performs well on images of people which exhibit strong variation in appearance as well as articulations. Finally, we show generalization performance by cross-dataset evaluation.

\section{Related Work}
The idea of representing articulated objects in general, and human pose in particular, as a graph of parts has been advocated from the early days of computer vision \cite{nevatia1977description}. The so called Pictorial Strictures (PSs), introduced by Fishler and Elschlager~\cite{fischler1973representation}, were made tractable and practical by Felzenszwalb and Huttenlocher~\cite{felzenszwalb2005pictorial} using the distance transform trick. As a result, a wide variety of PS-based models with practical significance were subsequently developed.

The above tractability, however, comes with the limitation of having a tree-based pose models with simple binary potential not depending on image data. As a result, research has focused on enriching the representational power of the models while maintaining tractability. Earlier attempts to achieve this were based on richer part detectors \cite{ramanan2006learning,andriluka2009pictorial,eichner2009better}. More recently, a wide variety of models expressing complex joint relationships were proposed. Yang and Ramanan~\cite{yang2011articulated} use a mixture model of parts. Mixture models on the full model scale, by having mixture of PSs, have been studied by Johnson and Everingham~\cite{Johnson11}. Richer higher-order spatial relationships were captured in a hierarchical model by Tian et al.~\cite{tian2012exploring}. A different approach to capture higher-order relationship is through image-dependent PS models, which can be estimated via a global classifier \cite{wang2013beyond,modec13,pishchulin2013poselet}.

Approaches which ascribe to our philosophy of reasoning about pose in a holistic manner have shown limited practicality. Mori and Malik~\cite{mori2002estimating} try to find for each test image the closest exemplar from a set of labeled images and transfer the joint locations. A similar nearest neighbor setup is employed by Shakhnarovich et al.~\cite{shakhnarovich2003fast}, who however use locality sensitive hashing. More recently, Gkioxari et al.~\cite{gkioxariarticulated} propose a semi-global classifier for part configuration. This formulation has shown very good results on real-world data, however, it is based on linear classifiers with less expressive representation than ours and is tested on arms only. Finally, the idea of pose regression has been employed by Ionescu et al.~\cite{ionescu2011latent}, however they reason about 3D pose.

The closest work to ours uses convolution NNs together with Neighborhood Component Analysis to regress toward a point in an embedding representing pose \cite{taylor2010pose}. However, this work does not employ a cascade of networks. Cascades of DNN regressors have been used for localization, however of facial points \cite{sun2013deep}. On the related problem of face pose estimation, Osadchy et al.~\cite{osadchy2007synergistic} employ a NN-based pose embedding trained with a contrastive loss.
\section{Deep Learning Model for Pose Estimation}
We use the following notation. To express a pose, we encode the locations of all $k$ body joints in \textit{pose vector} defined as $\mathbf{y}=(\ldots, \mathbf{y}_i^T,\ldots )^T, i\in\{1,\ldots, k\}$, where $\mathbf{y}_i$ contains the $x$ and $y$ coordinates of the $i^\text{th}$ joint. A labeled image is denoted by $(x,\mathbf{y})$ where $x$ stands for the image data and $\mathbf{y}$ is the ground truth pose vector.

Further, since the joint coordinates are in absolute image coordinates, it proves beneficial to normalize them w.~r.~t.~a box $b$ bounding the human body or parts of it. In a trivial case, the box can denote the full image. Such a box is defined by its center $b_c\in\mathbb{R}^2$ as well as width $b_w$ and height $b_h$: $b = (b_c,b_w,b_h)$. Then the joint $\mathbf{y}_i$ can be translated by the box center and scaled by the box size which we refer to as normalization by $b$:
\begin{equation}\label{eq:normalization}
N(\mathbf{y}_i; b) = \left(\begin{array}{ccc} 1/b_w & 0 \\ 0 & 1/b_h \end{array}\right)(\mathbf{y}_i - b_c)
\end{equation}
Further, we can apply the same normalization to the elements of pose vector $N(\mathbf{y}; b)=(\ldots, N(\mathbf{y}_i; b)^T,\ldots )^T$ resulting in a \textit{normalized pose vector}. Finally, with a slight abuse of notation, we use $N(x; b)$ to denote a crop of the image $x$ by the bounding box b, which de facto normalizes the image by the box. For brevity we denote by $N(\cdot)$ normalization with $b$ being the full image box.


\subsection{Pose Estimation as DNN-based Regression}\label{sec:pose_regression}
\begin{figure*}[t]
{\centering
\includegraphics[width=0.99\textwidth]{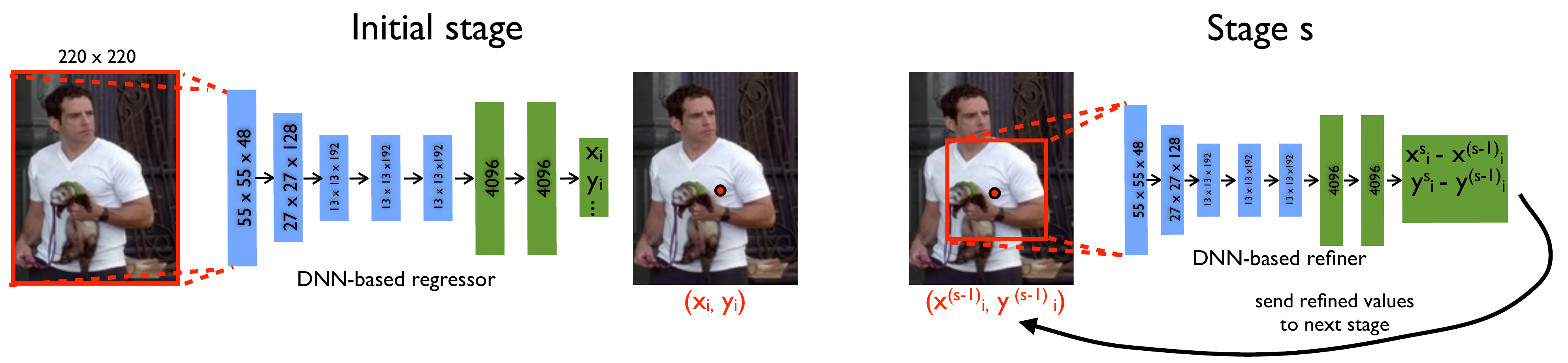}
\caption{\label{fig:intro_pose} Left: schematic view of the DNN-based pose regression. We visualize the network layers with their corresponding dimensions, where convolutional layers are in blue, while fully connected ones are in green. We do not show the parameter free layers. Right: at stage $s$, a refining regressor is applied on a sub image to refine a prediction from the previous stage.}
}
\end{figure*}
In this work, we treat the problem of pose estimation as regression, where the we train and use a function $\psi(x;\theta)\in \mathbb{R}^{2k}$ which for an image $x$ regresses to a normalized pose vector, where $\theta$ denotes the parameters of the model.  Thus, using the normalization transformation from Eq.~(\ref{eq:normalization}) the pose prediction $y^*$ in absolute image coordinates reads
\begin{equation}\label{eq:pose_inference}
y^* = N^{-1}(\psi(N(x); \theta))
\end{equation}

Despite its simple formulation, the power and complexity of the method is in $\psi$, which is based on a convolutional Deep Neural Network (DNN). Such a convolutional network consists of several layers -- each being a linear transformation followed by a non-linear one. The first layer takes as input an image of predefined size and has a size equal to the number of pixels times three color channels. The last layer outputs the target values of the regression, in our case $2k$ joint coordinates.

We base the architecture of the $\psi$ on the work by Krizhevsky et al.~\cite{krizhevsky2012imagenet} for image classification since it has shown outstanding results on object localization as well \cite{szegedy2013object}. In a nutshell, the network consists of $7$ layers (see Fig.~\ref{fig:intro_pose} left). Denote by $C$ a convolutional layer, by $LRN$ a local response normalization layer, $P$ a pooling layer and by $F$ a fully connected layer. Only $C$ and $F$ layers contain learnable parameters, while the rest are parameter free. Both $C$ and $F$ layers consist of a linear transformation followed by a nonlinear one, which in our case is a rectified linear unit.  For $C$ layers, the size is defined as $\text{width} \times \text{height} \times \text{depth}$, where the first two dimensions have a spatial meaning while the depth defines the number of filters. If we write the size of each layer in parentheses, then the network can be described concisely as $C(55\times 55 \times 96) - LRN - P - C(27 \times 27 \times 256) - LRN - P - C(13 \times 13 \times 384) - C(13 \times 13 \times 384) - C(13 \times 13 \times 256) - P - F(4096) - F(4096)$. The filter size for the first two $C$ layers is $11\times 11$ and $5\times 5$ and for the remaining three is $3\times 3$. Pooling is applied after three layers and contributes to increased performance despite the reduction of resolution. The input to the net is an image of $220 \times 220$ which via stride of $4$ is fed into the network. The total number of parameters in the above model is about 40M. For further details, we refer the reader to \cite{krizhevsky2012imagenet}. 

The use of a generic DNN architecture is motivated by its outstanding results on both classification and localization problems. In the experimental section we show that such a generic architecture can be used to learn a model resulting in state-of-art or better performance on pose estimation as well. Further, such a model is a truly holistic one --- the final joint location estimate is based on a complex nonlinear transformation of the full image. 

Additionally, the use of a DNN obviates the need to design a domain specific pose model. Instead such a model and the features are learned from the data. Although the regression loss does not model explicit interactions between joints, such are implicitly captured by all of the $7$ hidden layers -- all the internal features are shared by all joint regressors.
\vspace{-0.5cm}
\paragraph{Training} The difference to \cite{krizhevsky2012imagenet} is the loss. Instead of a classification loss, we train a linear regression on top of the last network layer to predict a pose vector by minimizing $L_2$ distance between the prediction and the true pose vector. Since the ground truth pose vector is defined in absolute image coordinates and poses vary in size from image to image, we normalize our training set $D$ using the normalization from Eq.~(\ref{eq:normalization}):
\begin{equation}\label{eq:normalized_training_set}
D_N = \{(N(x), N(\mathbf{y})) | (x, \mathbf{y})\in D\}
\end{equation}
Then the $L_2$ loss for obtaining optimal network parameters reads:
\begin{equation}\label{eq:learning_objective}
\arg\min_\theta\sum_{(x,y) \in D_N} \sum_{i=1}^k||\mathbf{y}_i - \psi_i(x;\theta)||^2_2
\end{equation}
For clarity we write out the optimization over individual joints. It should be noted, that the above objective can be used even if for some images not all joints are labeled. In this case, the corresponding terms in the sum would be omitted. 

The above parameters $\theta$ are optimized for using Backpropagation in a distributed online implementation. For each mini-batch of size $128$, adaptive gradient updates are computed \cite{duchi2010adagrad}. The learning rate, as the most important parameter, is set to $0.0005$. Since the model has large number of parameters and the used datasets are of relatively small size, we augment the data using large number of randomly translated image crops (see Sec.~\ref{sec:cascade}), left/right flips as well as DropOut regularization for the $F$ layers set to $0.6$.


\subsection{Cascade of Pose Regressors}\label{sec:cascade}
The pose formulation from the previous section has the advantage that the joint estimation is based on the full image and thus relies on context. However, due to its fixed input size of $220\times 220$, the network has limited capacity to look at detail -- it learns filters capturing pose properties at coarse scale. These are necessary to estimate rough pose but insufficient to always precisely localize the body joints. Note that we cannot easily increase the input size since this will increase the already large number of parameters. In order to achieve better precision, we propose to train a cascade of pose regressors. At the first stage, the cascade starts off by estimating an initial pose as outlined in the previous section. At subsequent stages, additional DNN regressors are trained to predict a displacement of the joint locations from previous stage to the true location. Thus, each subsequent stage can be thought of as a refinement of the currently predicted pose, as shown in Fig.~\ref{fig:intro_pose}.

Further, each subsequent stage uses the predicted joint locations to focus on the relevant parts of the image -- sub-images are cropped  around the predicted joint location from previous stage and the pose displacement regressor for this joint is applied on this sub-image. In this way, subsequent pose regressors see higher resolution images and thus learn features for finer scales which ultimately leads to higher precision. 

 We use the same network architecture for all stages of the cascade but learn different network parameters. For stage $s\in\{1, \ldots, S\}$ of total $S$ cascade stages, we denote by $\theta_s$ the learned network parameters. Thus, the pose displacement regressor reads $\psi(x; \theta_s)$. To refine a given joint location $\mathbf{y}_i$ we will consider a \textit{joint bounding box} $b_i$ capturing the sub-image around $\mathbf{y}_i$: $b_i(\mathbf{y};\sigma) = (\mathbf{y}_i, \sigma\text{diam}(\mathbf{y}), \sigma\text{diam}(\mathbf{y}))$ having as center the $i$-th joint and as dimension the pose diameter scaled by $\sigma$. The diameter $\text{diam}(\mathbf{y})$ of the pose is defined as the distance between opposing joints on the human torso, such as left shoulder and right hip, and depends on the concrete pose definition and dataset.
  
 Using the above notation, at the stage $s = 1$ we start with a bounding box $b^0$ which either encloses the full image or is obtained by a person detector. We obtain an initial pose: 
\begin{eqnarray}
&\text{Stage $1$}:& \textbf{y}^1 \leftarrow  N^{-1}(\psi(N(x; b^0);\theta_1); b^0) 
\end{eqnarray} 
 At each subsequent stage $s\geq 2$, for all joints  $i\in\{1,\ldots, k\}$ we regress first towards a refinement displacement $\textbf{y}^s_i - \textbf{y}^{(s-1)}_i$ by applying a regressor on the sub image defined by $b_i^{(s-1)}$ from previous stage $(s-1)$. Then, we estimate new joint boxes $b_i^s$:
 \begin{eqnarray}
&\text{Stage $s$:}& \textbf{y}^s_i \leftarrow  \textbf{y}^{(s-1)}_i + N^{-1}(\psi_i(N(x;b); \theta_s);b)\\
&& \quad \text{ for } b = b_i^{(s-1)} \nonumber \\
&&  b_i^s \leftarrow (\textbf{y}_i^s, \sigma\text{diam}(\textbf{y}^s), \sigma\text{diam}(\textbf{y}^s))
\end{eqnarray}


We apply the cascade for a fixed number of stages $S$, which is determined as explained in Sec.~\ref{sec:experimental_setup}.
  \vspace{-0.5cm}
 \paragraph{Training} The network parameters $\theta_1$ are trained as outlined in Sec.~\ref{sec:pose_regression}, Eq.~(\ref{eq:learning_objective}). At subsequent stages $s\geq 2$, the training is done identically with one important difference. Each joint $i$ from a training example $(x,\mathbf{y})$ is normalized using a different bounding box $(\mathbf{y}^{(s-1)}_i, \sigma\text{diam}(\mathbf{y}^{(s-1)}), \sigma\text{diam}(\mathbf{y}^{(s-1)}))$ -- the one centered at the prediction for the same joint obtained from previous stage -- so that we condition the training of the stage based on the model from previous stage.
 
Since deep learning methods have large capacity, we augment the training data by using multiple normalizations for each image and joint. Instead of using the prediction from previous stage only, we generate simulated predictions. This is done by randomly displacing the ground truth location for joint $i$ by a vector sampled at random from a 2-dimensional Normal distribution $\mathcal{N}_i^{(s-1)}$ with mean and variance equal to the mean and variance of the observed  displacements $(\mathbf{y}_i^{(s-1)} - \mathbf{y}_i)$ across all examples in the training data. The full augmented training data can be defined by first sampling an example and a joint from the original data at uniform and then generating a simulated prediction based on a sampled displacement $\delta$ from $\mathcal{N}_i^{(s-1)}$:
\begin{eqnarray}
D^{s}_A&=&\{(N(x;b), N(\mathbf{y}_i; b))|\nonumber\\
&&(x,\mathbf{y}_i)\sim D, \delta\sim\mathcal{N}^{(s-1)}_i, \nonumber\\
&&b = (\mathbf{y}_i+\delta, \sigma\text{diam}(\mathbf{y}))\}\nonumber
\end{eqnarray}

The training objective for cascade stage $s$ is done as in Eq.~(\ref{eq:learning_objective}) by taking extra care to use the correct normalization for each joint:
\begin{equation}\label{eq:cascade_learning_objective}
\theta_s = \arg\min_\theta\sum_{(x,\mathbf{y}_i) \in D^{s}_A} ||\mathbf{y}_i - \psi_i(x;\theta)||^2_2
\end{equation}
 
%
\section{Empirical Evaluation}
\subsection{Setup}\label{sec:experimental_setup}
\paragraph{Datasets} There is a wide variety of benchmarks for human pose estimation. In this work we use datasets, which have large number of training examples sufficient to train a large model such as the proposed DNN, as well as are realistic and challenging.

The first dataset we use is Frames Labeled In Cinema (FLIC), introduced by \cite{modec13}, which consists of 4000 training and 1000 test images obtained from popular Hollywood movies. The images contain people in diverse poses and especially diverse clothing. For each labeled human, $10$ upper body joints are labeled. 

The second dataset we use is Leeds Sports Dataset \cite{Johnson10} and its extension \cite{Johnson11}, which we will jointly denote by LSP. Combined they contain $11000$ training and $1000$ testing images. These are images from sports activities and as such are quite challenging in terms of appearance and especially articulations. In addition, the majority of people have 150 pixel height which makes the pose estimation even more challenging. In this dataset, for each person the full body is labeled with total $14$ joints.

For all of the above datasets, we define the diameter of a pose $\mathbf{y}$ to be the distance between a shoulder and hip from opposing sides and denote it by $\text{diam}(\mathbf{y})$. It should be noted, that the joints in all datasets are arranged in a tree kinematically mimicking the human body. This allows for a definition of a limb being a pair of neighboring joints in the pose tree.
\vspace{-0.5cm}
\paragraph{Metrics} In order to be able to compare with published results we will use two widely accepted evaluation metrics. Percentage of Correct Parts (PCP) measures detection rate of limbs, where a limb is considered detected if the distance between the two predicted joint locations and the true limb joint locations is at most half of the limb length \cite{eichner2010articulated}. PCP was the initially preferred metric for evaluation, however it has the drawback of penalizing shorter limbs, such as lower arms, which are usually harder to detect.

To address this drawback, recently detection rates of joints are being reported using a different detection criterion -- a joint is considered detected if the distance between the predicted and the true joint is within a certain fraction of the torso diameter. By varying this fraction, detection rates are obtained for varying degrees of localization precision. This metric alleviates the drawback of PCP since the detection criteria for all joints are based on the same distance threshold. We refer to this metric as Percent of Detected Joints (PDJ). 
\vspace{-0.5cm}
\paragraph{Experimental Details}
For all the experiments we use the same network architecture. Inspired by \cite{Ferrari08}, we use a body detector on FLIC to obtain initially a rough estimate of the human body bounding box. It is based on a face detector -- the detected face rectangle is enlarged by a fixed scaler. This scaler is determined on the training data such that it contains all labeled joints. This face-based body detector results in a rough estimate, which however presents a good starting point for our approach. For LSP we use the full image as initial bounding box since the humans are relatively tightly cropped by design.

Using a small held-out set of 50 images for both datasets to determine the algorithm hyperparameters. To measure optimality of the parameters we used average over PDJ at $0.2$ across all joints. The scaler $\sigma$, which defines the size of the refinement joint bounding box as a fraction of the pose size, is determined as follows: for FLIC we chose $\sigma = 1.0$ after exploring values $\{0.8,1.0,1.2\}$, for LSP we use $\sigma = 2.0$ after trying $\{1.5,1.7,2.0,2.3\}$. The number of cascade stages $S$ is determined by training stages until the algorithm stopped improving on the held-out set. For both FLIC and LSP we arrived at $S=3$.

To improve generalization, for each cascade stage starting at $s=2$ we augment the training data by sampling $40$ randomly translated crop boxes for each joint as explained in Sec.~\ref{sec:cascade}. Thus, for LSP with $14$ joints and after mirroring the images and sampling the number training examples is $11000 \times 40 \times 2 \times 14 = 12M$, which is essential for training a large network as ours.

The presented algorithm allows for an efficient implementation.  The running time is approx.~$0.1s$ per image, as measured on a $12$ core CPU. This compares favorably to other approaches, as some of the current state-of-art approaches have higher complexity: \cite{modec13} runs in approx.~$4s$, while \cite{yang2011articulated} runs in $1.5s$. The training complexity, however, is higher. The initial stage was trained within $3$ days on approx.~$100$ workers, most of the final performance was achieved after $12$ hours though. Each refinement stage was trained for $7$ days since the amount of data was $40\times$ larger than the one for the initial stage due to the data augmentation in Sec.~\ref{sec:cascade}. Note that using more data led to increased performance. 
\begin{figure*}[t]
{\centering
\includegraphics[width=0.35\textwidth]{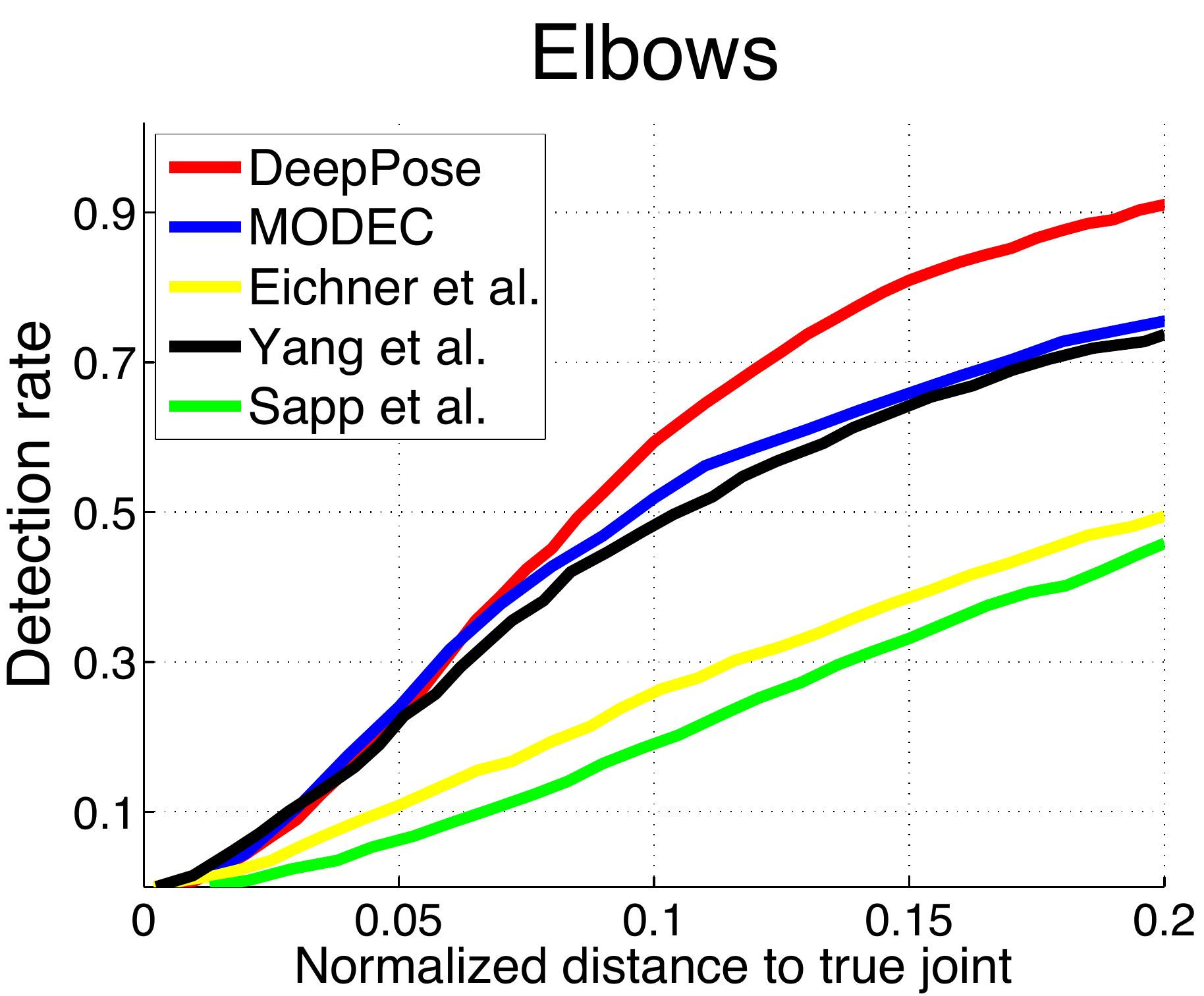}
\includegraphics[width=0.35\textwidth]{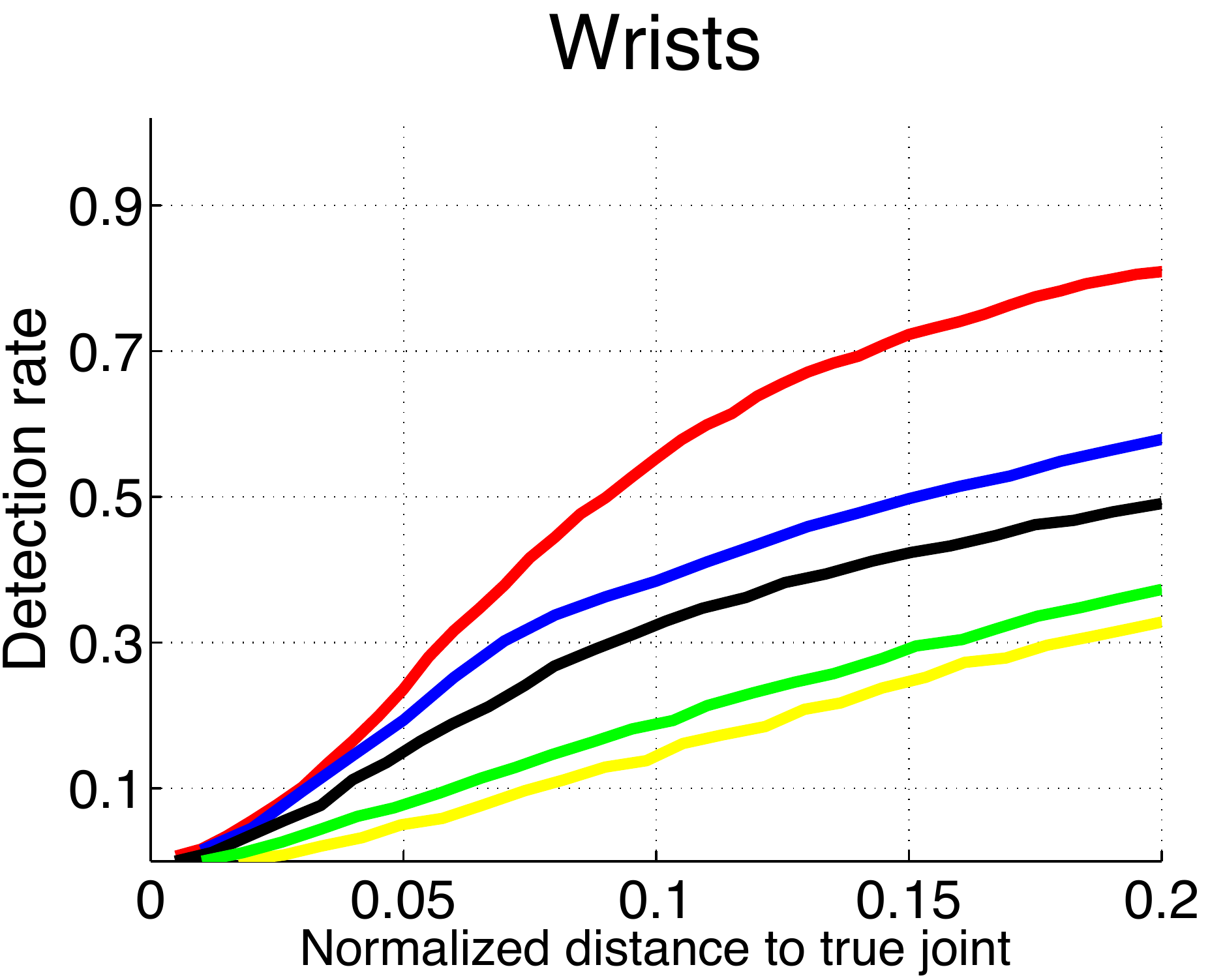}
\caption{\label{fig:results_flic} Percentage of detected joints (PDJ) on FLIC for two joints: elbow and wrist. We compare DeepPose, after two cascade stages, with four other approaches.}
}
\end{figure*}
\subsection{Results and Discussion}
\paragraph{Comparisons}
We present comparative results to other approaches. We compare on LSP using PCP metric in Fig.~\ref{tab:pcp_results}. We show results for the four most challenging limbs -- lower and upper arms and legs -- as well as the average value across these limbs for all compared algorithms. We clearly outperform all other approaches, especially achieving better estimation for legs. For example, for upper legs we obtain $0.78$ up from $0.74$ for the next best performing method. It is worth noting that while the other approaches exhibit strengths for particular limbs, none of the other dataset consistently dominates across all limbs. In contrary, DeepPose shows strong results for all challenging limbs.

\begin{figure}
{\centering
\includegraphics[width=0.23\textwidth]{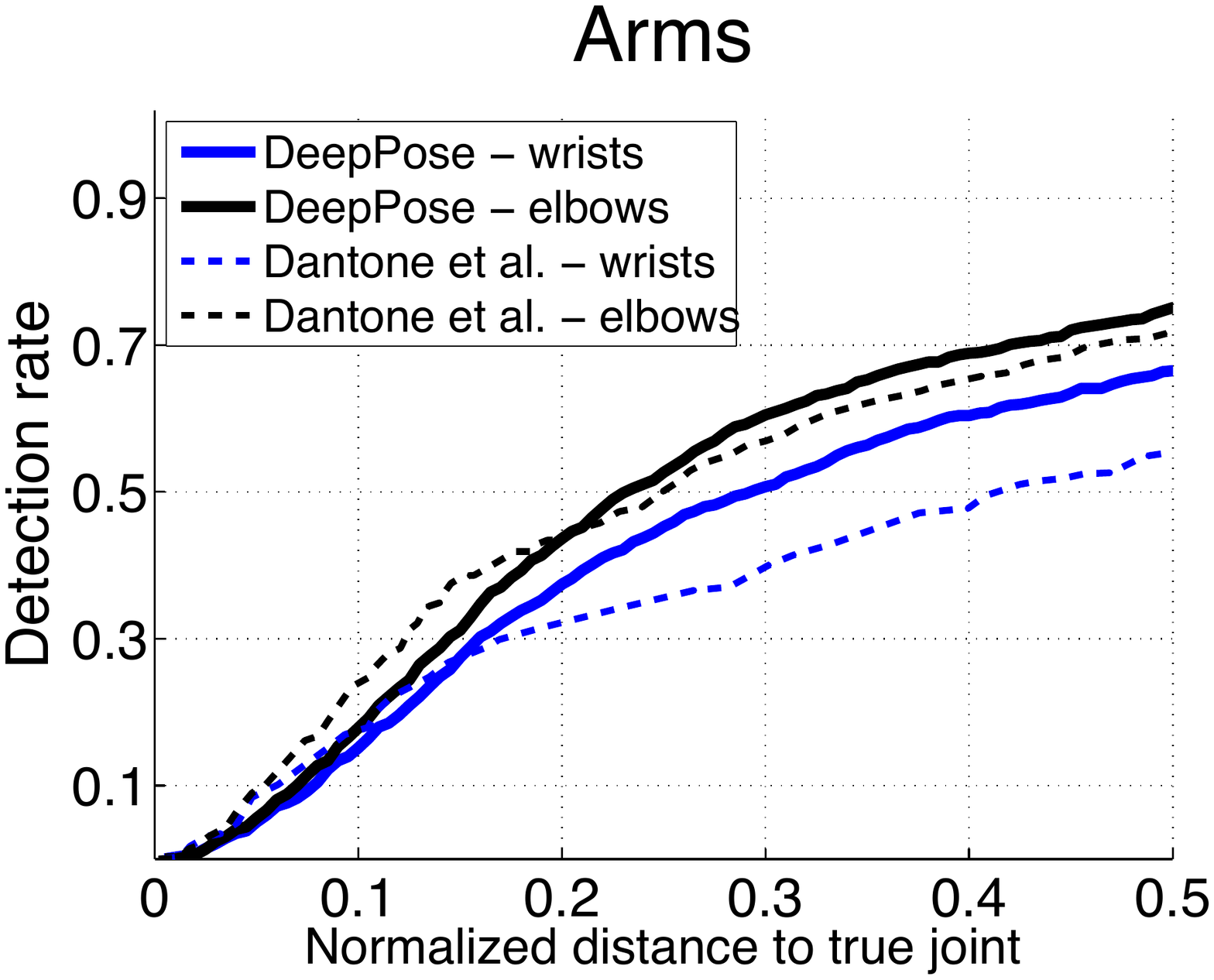}
\includegraphics[width=0.23\textwidth]{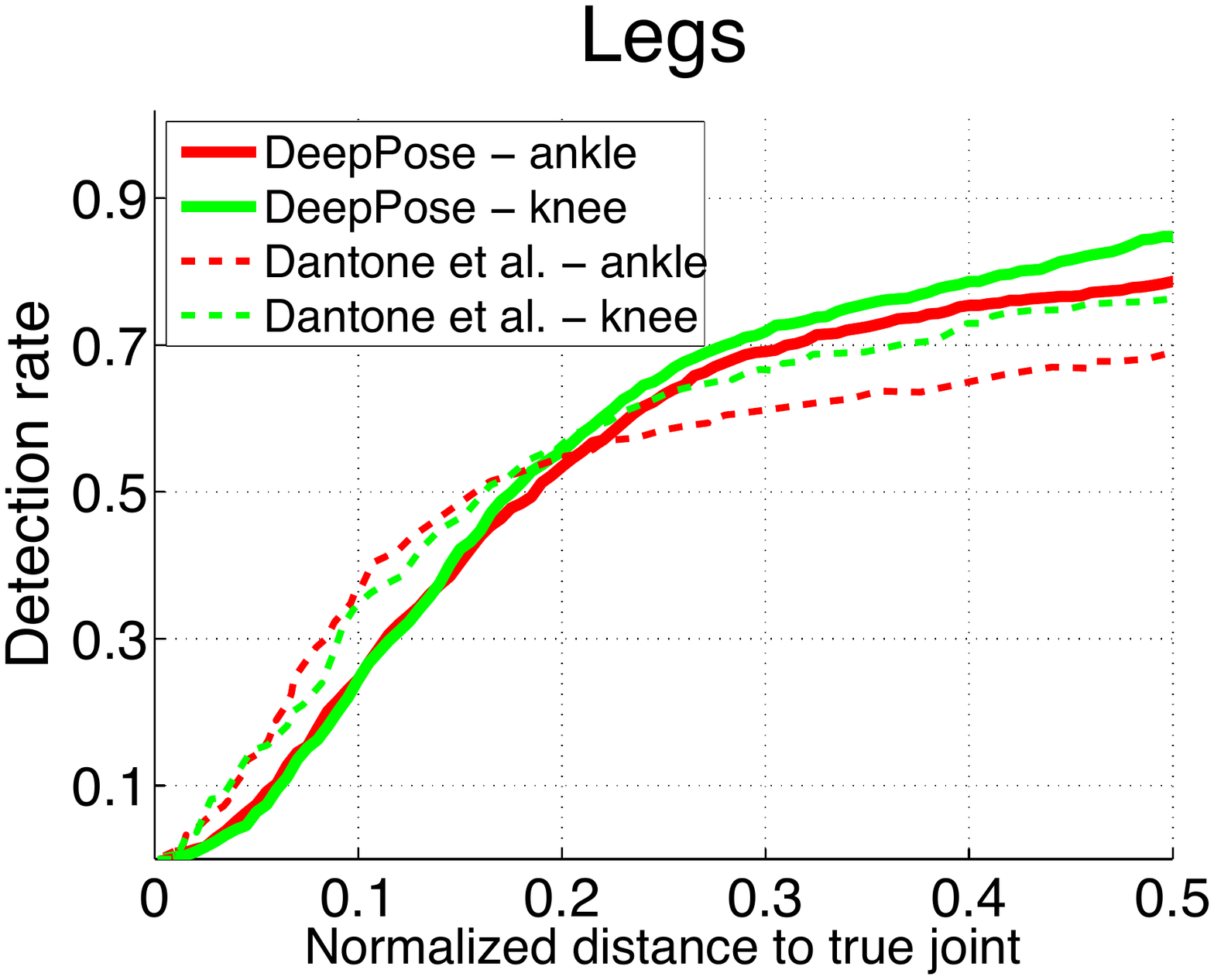}
\caption{\label{fig:pdj_lsp} Percentage of detected joints (PDJ) on LSP for four limbs for DeepPose and Dantone et al.~\cite{dantone13regressors} over an extended range of distances to true joint: $[0, 0.5]$ of the torso diameter. Results of DeepPose are plotted with solid lines while all the results by \cite{dantone13regressors} are plotted in dashed lines. Results for the same joint from both algorithms are colored with same color.}
}
\end{figure}

Using the PDJ metric allows us to vary the threshold for the distance between prediction and ground truth, which defines a detection. This threshold can be thought of as a localization precision at which detection rates are plotted. Thus one could compare approaches across different desired precisions. We present results on FLIC in Fig.~\ref{fig:results_flic} comparing against additional four methods as well is on LSP in Fig.~\ref{fig:pdj_lsp}. For each dataset we train and test according the protocol for each dataset. Similarly to previous experiment we outperform all five algorithms. Our gains are bigger in the low precision domain, in the cases where we detect rough pose without precisely localizing the joints. On FLIC, at normalized distance $0.2$ we obtain a an increase of detection rates by $0.15$ and $0.2$ for elbow and wrists against the next best performing method. On LSP, at normalized distance $0.5$ we get an absolute increase of $0.1$. At low precision regime of normalized distance of $0.2$ for LSP we show comparable performance for legs and slightly worse arms. This can be attributed to the fact that the DNN-based approach computes joint coordinates using 7 layers of transformation, some of which contain max pooling.

Another observation is that our approach works well for both appearance heavy movie data as well as string articulation such as the sports images in LSP.
\begin{table}	
{\centering
{\small
\begin{tabular}{| c || c | c | c | c || c |} 
\hline
\multirow{2}{*}{Method} & \multicolumn{2}{|c|} {Arm} & \multicolumn{2}{|c||} {Leg} & \multirow{2}{*}{Ave.}\\
\cline{2-5}
 & Upper & Lower & Upper & Lower & \\
\hline\hline
DeepPose-st1& 0.5 & 0.27 & 0.74 & 0.65 & 0.54\\
\hline
DeepPose-st2& \textbf{0.56} & 0.36 & \textbf{0.78} & 0.70& 0.60\\
\hline
DeepPose-st3& \textbf{0.56} & \textbf{0.38} & 0.77 & \textbf{0.71} & \textbf{0.61}\\
\hline\hline
Dantone et al. \cite{dantone13regressors} & 0.45 & 0.25 & 0.65 & 0.61 & 0.49 \\
\hline
Tian et al. \cite{tian2012exploring}$^*$ & 0.52 & 0.33 & 0.70 & 0.60 &  0.56\\
\hline
Johnson et al. \cite{Johnson11} & 0.54 & \textbf{0.38} & 0.75 & 0.66 & 0.58\\
\hline
Wang et al. \cite{wang2013beyond}$^*$ & \textbf{0.565} & 0.37 & 0.76 & 0.68 & 0.59 \\ 
\hline 
Pishchulin \cite{pishchulin2013poselet}$^\circ$ & 0.49 & 0.32 & 0.74 & 0.70 & 0.56 \\
\hline 
\end{tabular} } }
\caption{\label{tab:pcp_results}Percentage of Correct Parts (PCP) at 0.5 on LSP for DeepPose as well as five state-of-art approaches. $^*$The authors use a slightly looser version of PCP in which for each limb the average distance from predicted limb joints belonging to the corresponding true joints is used to determine whether the limb is correctly detected. $^\circ$The authors use person-centric joint annotations.}
\end{table} 

\begin{figure}
{\centering
\includegraphics[width=0.22\textwidth]{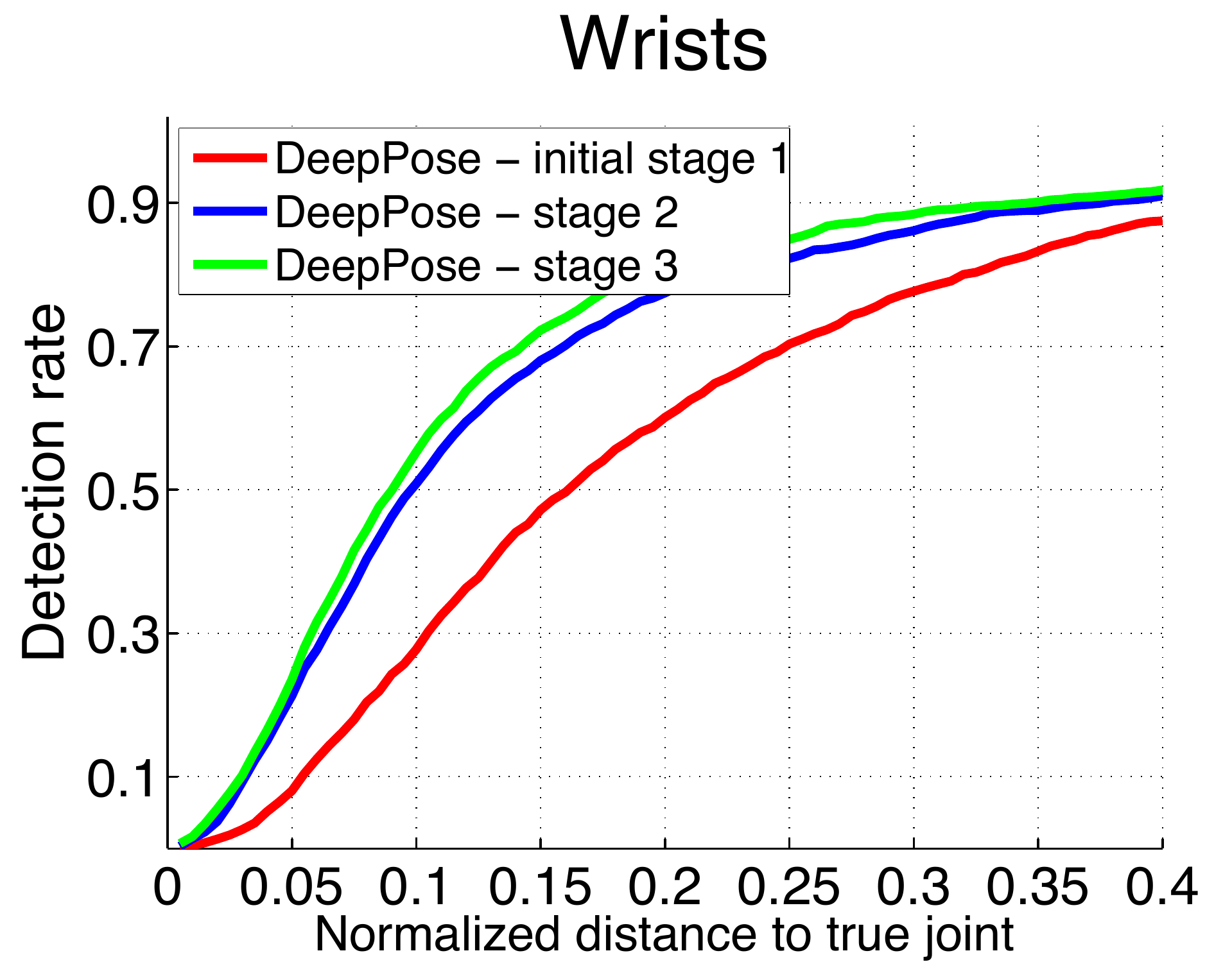}
\includegraphics[width=0.22\textwidth]{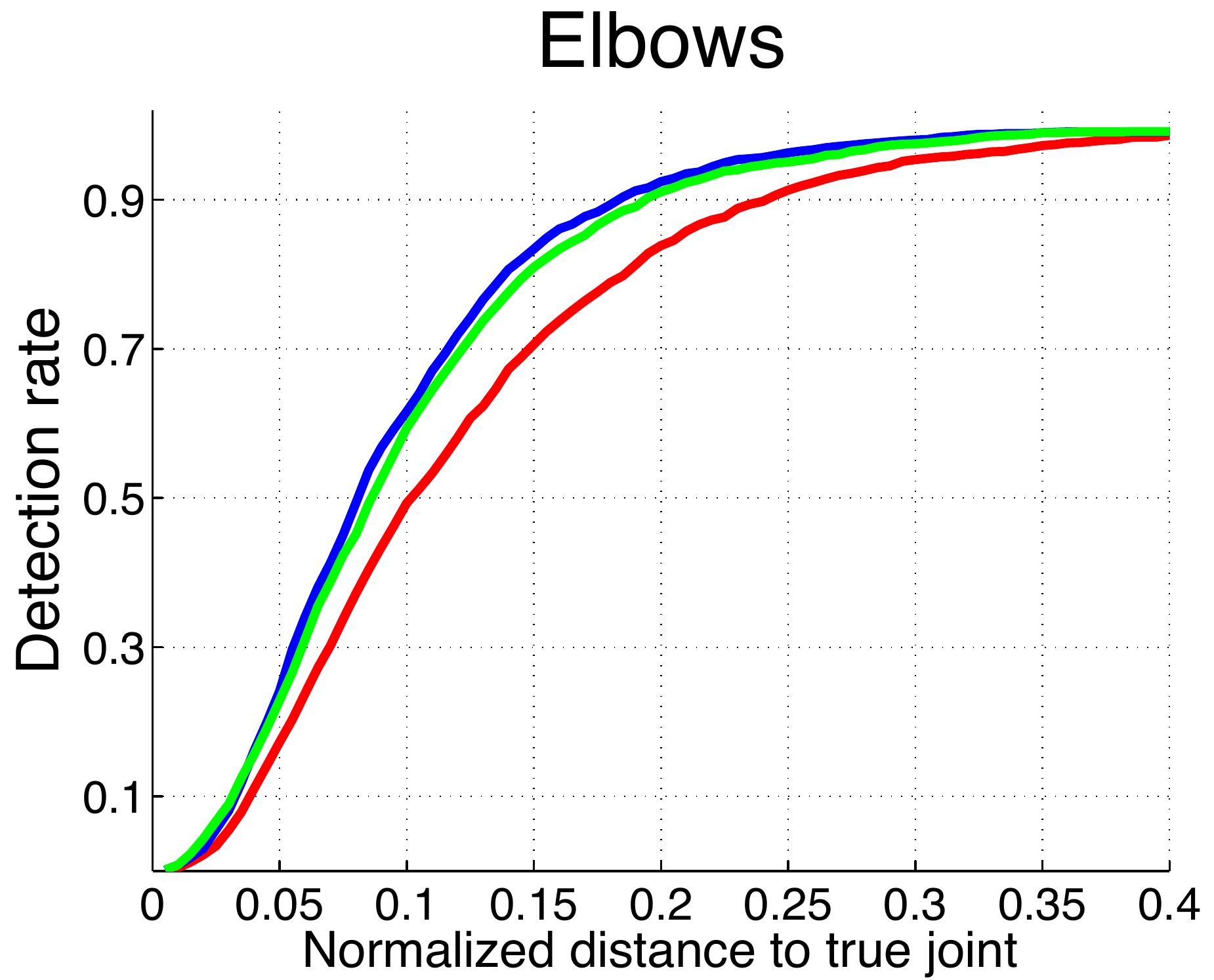}
\caption{\label{fig:cascade_evaluation}Percent of detected joints (PDJ) on FLIC or the first three stages of the DNN cascade. We present results over larger spectrum of normalized distances between prediction and ground truth.}
}
\end{figure}

\vspace{-0.5cm}
\paragraph{Effects of cascade-based refinement} A single DNN-based joint regressor gives rough joint location. However, to obtain higher precision the subsequent stages of the cascade, which serve as a refinement of the initial prediction, are of paramount importance. To see this, in Fig.~\ref{fig:cascade_evaluation} we present the joint detections at different precisions for the initial prediction as well as two subsequent cascade stages. As expected, we can see that the major gains of the refinement procedure are at high-precision regime of at normalized distances of $[0.15, 0.2]$.  Further, the major gains are achieved after one stage of refinement. The reason being that subsequent stages end up using smaller sub-images around each joint. And although the subsequent stages look at higher resolution inputs, they have more limited context.

\begin{figure}
{\centering
\includegraphics[width=0.4\textwidth]{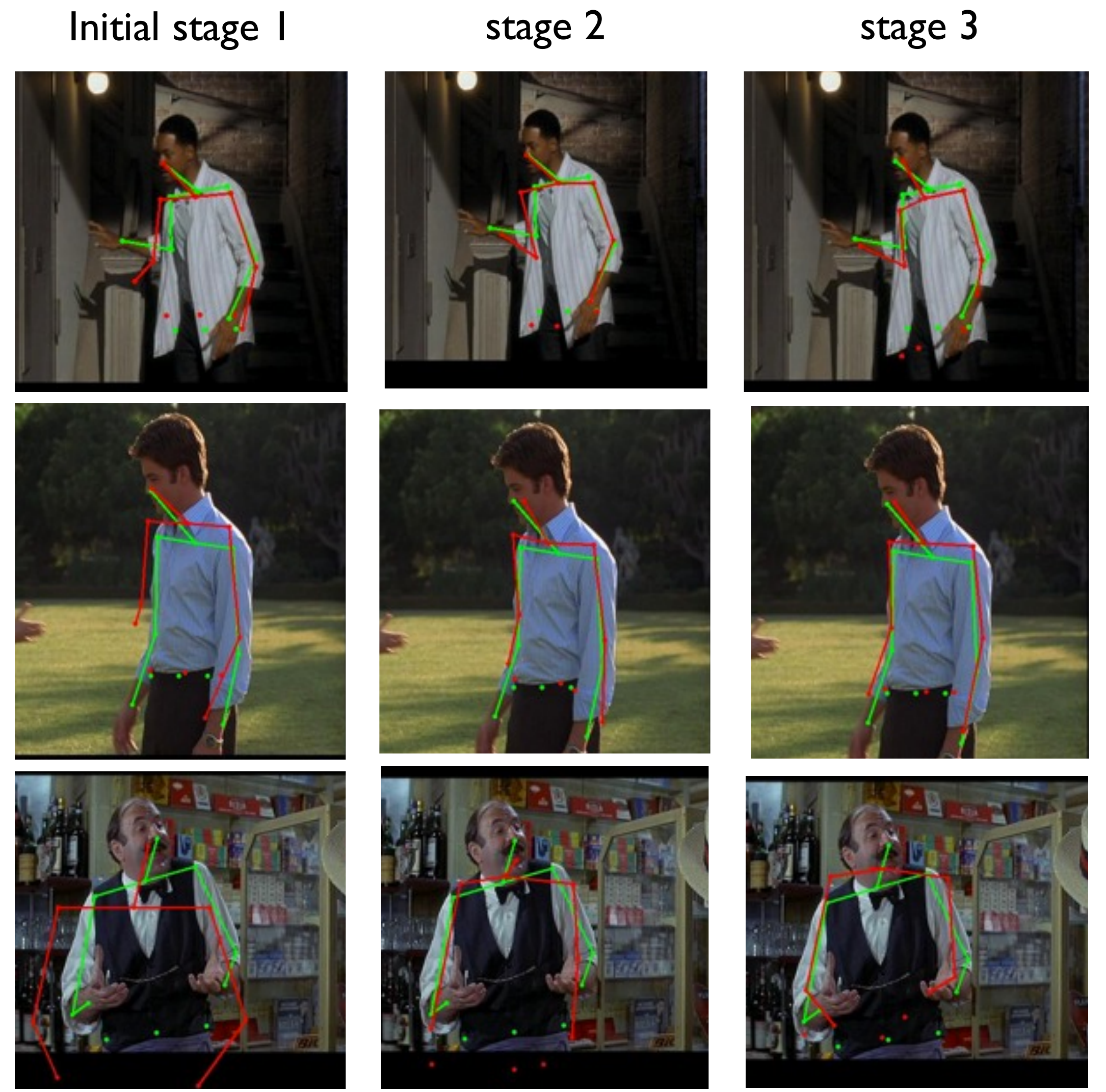}
\caption{\label{fig:cascade_examples}Predicted poses in red and ground truth poses in green for the first three stages of a cascade for three examples. }
}\end{figure}

Examples of cases, where refinement helps, are visualized in Fig.~\ref{fig:cascade_examples}. The initial stage is usually successful at estimating a roughly correct pose, however, this pose is not "snapped" to the correct one. For example, in row three the pose has the right shape but incorrect scale. In the second row, the predicted pose is translated north from the ideal one. In most cases, the second stage of the cascade resolves this snapping problem and better aligns the joints. In more rare cases, such as in first row, further facade stages improve on individual joints. 

\vspace{-0.5cm}
\paragraph{Cross-dataset Generalization} To evaluate the generalization properties of our algorithm, we used the trained models on LSP and FLIC on two related datasets. The full-body model trained on LSP is tested on the test portion of the Image Parse dataset \cite{ramanan2006learning} with results presented in Table~\ref{tab:pcp_results_parse}. The ImageParse dataset is similar to LSP as it contains people doing sports, however it contains a lot of people from personal photo collections involved in other activities. Further, the upper-body model trained on FLIC was applied on the whole Buffy dataset \cite{Ferrari08}. We can see that our approach can retain state-of-art performance compared to other approaches. This shows good generalization abilities.

\begin{figure}
{\centering
\includegraphics[width=0.22\textwidth]{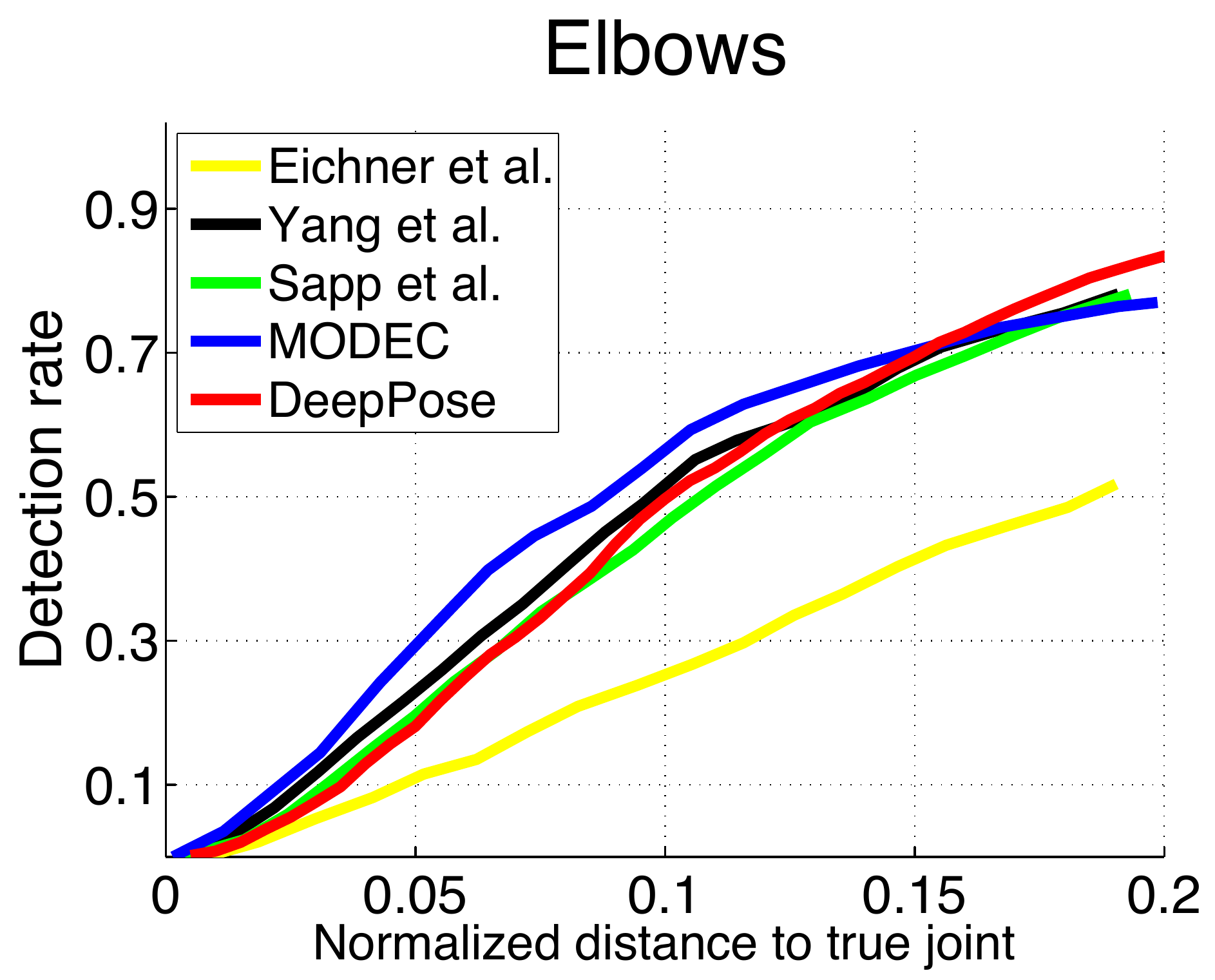}
\includegraphics[width=0.22\textwidth]{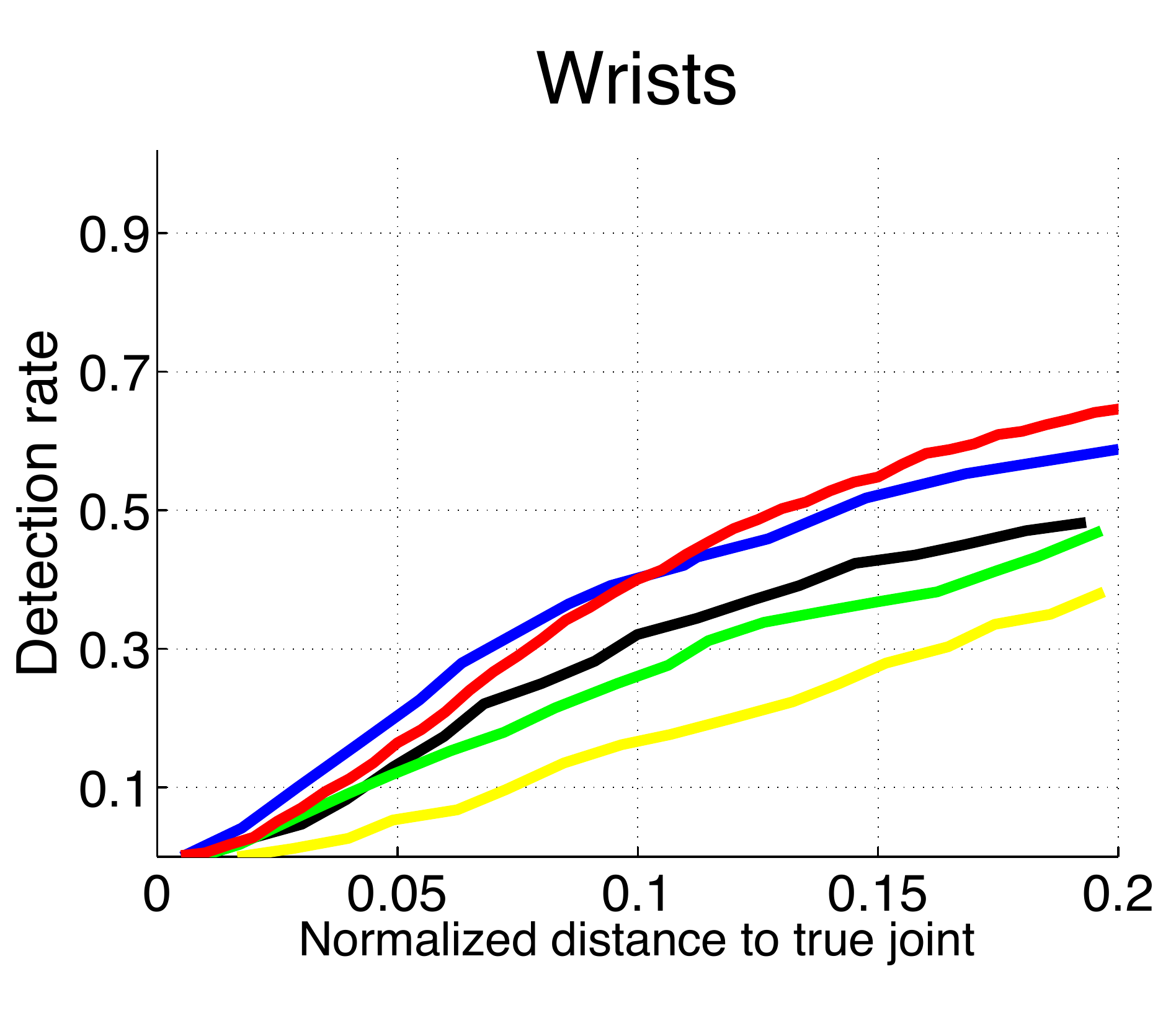}
\caption{\label{fig:pdj_buffy} Percentage of detected joints (PDJ) on Buffy dataset for two joints: elbow and wrist. The models have been trained on FLIC. We compare DeepPose, after two cascade stages, with four other approaches.}
}
\end{figure}

\begin{table}	
{\centering
{\small
\begin{tabular}{| c || c | c | c | c || c |} 
\hline
\multirow{2}{*}{Method} & \multicolumn{2}{|c|} {Arm} & \multicolumn{2}{|c||} {Leg} & \multirow{2}{*}{Ave.}\\
\cline{2-5}
 & Upper & Lower & Upper & Lower & \\
\hline
DeepPose& 0.8 & 0.75 & 0.71 & 0.5 & 0.69\\
\hline\hline
Pishchulin \cite{pishchulin2013poselet} & 0.80 & 0.70 & 0.59 & 037 &  0.62\\
\hline
Johnson et al. \cite{Johnson11} & 0.75 & 0.67 & 0.67 & 0.46 & 0.64\\
\hline
Yang et al. \cite{yang2011articulated} & 0.69 & 0.64 & 0.55 & 0.35 & 0.56\\
\hline
\end{tabular} } }
\caption{\label{tab:pcp_results_parse}Percentage of Correct Parts (PCP) at 0.5 on Image Parse dataset for DeepPose as well as two state-of-art approaches on Image Parse dataset. Results obtained from \cite{pishchulin2013poselet}.}
\end{table} 

\begin{figure*}
{\centering
\includegraphics[width=0.85\textwidth]{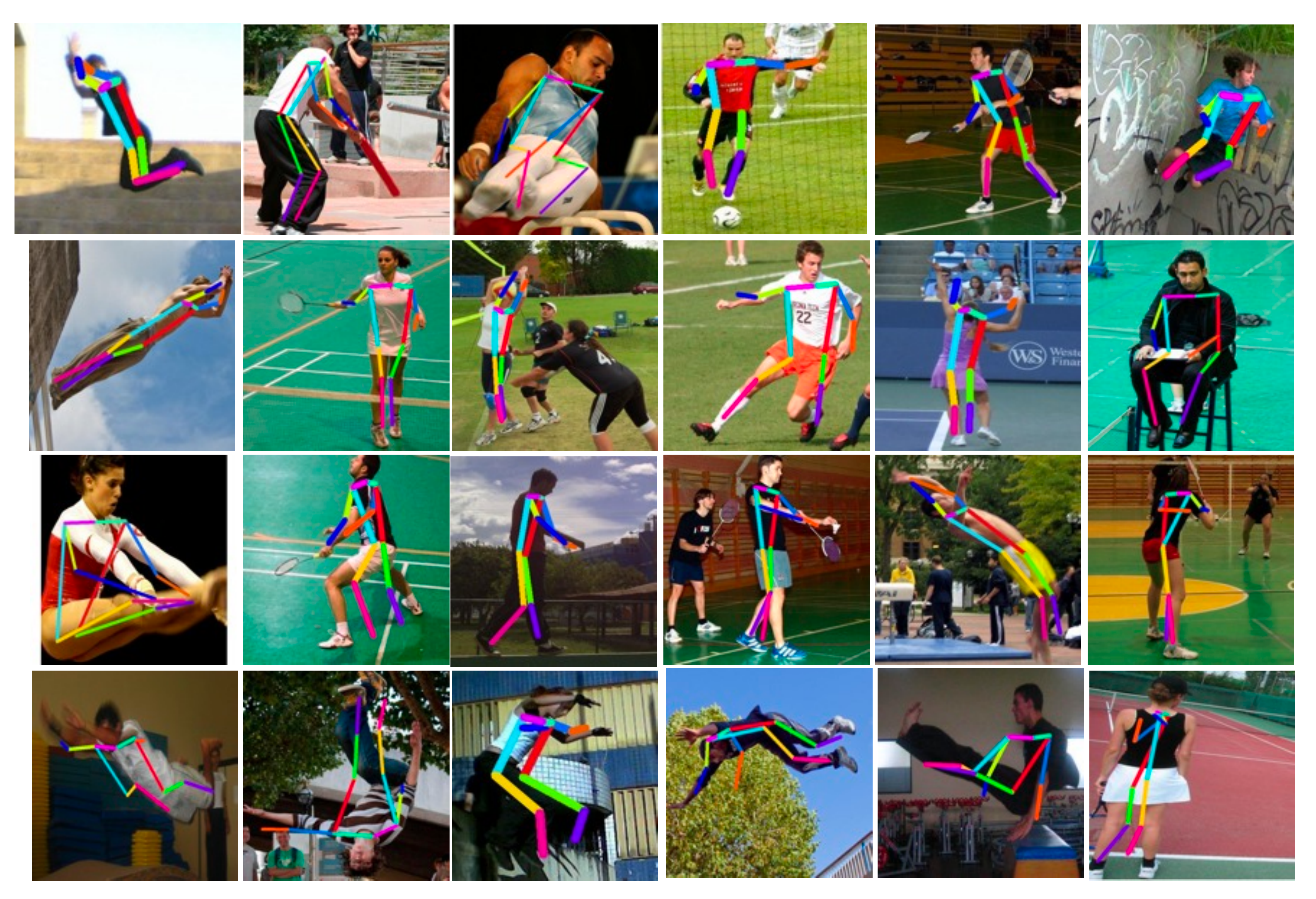}
\caption{\label{fig:examples_lsp}Visualization of pose results on images from LSP. Each pose is represented as a stick figure, inferred from predicted joints. Different limbs in the same image are colored differently, same limb across different images has the same color.}}
\end{figure*}

\begin{figure*}
{\centering
\includegraphics[width=0.85\textwidth]{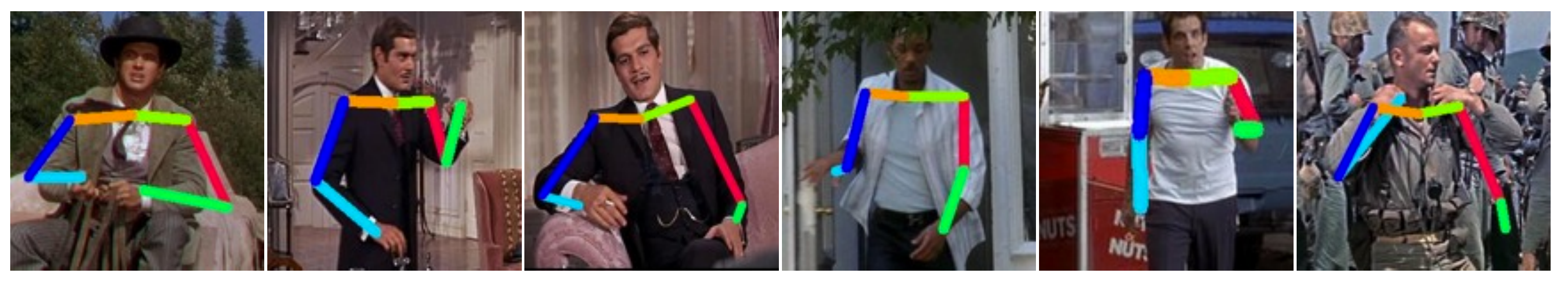}
\caption{\label{fig:examples_flic}Visualization of pose results on images from FLIC. Meaning of stick figures is the same as in Fig.~\ref{fig:examples_lsp} above.}}
\end{figure*}

\vspace{-0.5cm}
\paragraph{Example poses} To get a better idea of the performance of our algorithm, we visualize a sample of estimated poses on images from LSP in Fig.~\ref{fig:examples_lsp}. We can see that our algorithm is able to get correct pose for most of the joints under variety of conditions: upside-down people (row 1, column 1), severe foreshortening (row1, column 3), unusual poses (row 3, column 5), occluded limbs as the occluded arms in row 3, columns 2 and 6, unusual illumination conditions (row 3, column 3). In most of the cases, when the estimated pose is not precise, it still has a correct shape. For example, in the last row some of the predicted limbs are not aligned with the true locations, however the overall shape of the pose is correct. A common failure mode is confusing left with right side when the person was photographed from the back (row 6, column 6). Results on FLIC (see Fig.~\ref{fig:examples_flic}) are usually better with occasional visible mistakes on lower arms.

\section{Conclusion}
We present, to our knowledge, the first application of Deep Neural Networks (DNNs) to human pose estimation. Our formulation of the problem as DNN-based regression to joint coordinates and the presented cascade of such regressors has the advantage of capturing context and reasoning about pose in a holistic manner. As a result, we are able to achieve state-of-art or better results on several challenging academic datasets. 

Further, we show that using a generic convolutional neural network, which was originally designed for classification tasks, can be applied to the different task of localization. In future, we plan to investigate novel architectures which could be potentially better tailored towards localization problems in general, and in pose estimation in particular.
\vspace{-0.5cm}
\paragraph{Acknowledgements} I would like to thank Luca Bertelli, Ben Sapp and Tianli Yu for assistance with data and fruitful discussions.

{\small
\bibliographystyle{ieee}
\bibliography{egbib}

\begin{thebibliography}{10}\itemsep=-1pt

\bibitem{andriluka2009pictorial}
M.~Andriluka, S.~Roth, and B.~Schiele.
\newblock Pictorial structures revisited: People detection and articulated pose
  estimation.
\newblock In {\em CVPR}, 2009.

\bibitem{dantone13regressors}
M.~Dantone, J.~Gall, C.~Leistner, and L.~Van~Gool.
\newblock Human pose estimation using body parts dependent joint regressors.
\newblock In {\em CVPR}, 2013.

\bibitem{duchi2010adagrad}
J.~Duchi, E.~Hazan, and Y.~Singer.
\newblock Adaptive subgradient methods for online learning and stochastic
  optimization.
\newblock In {\em COLT}. ACL, 2010.

\bibitem{eichner2009better}
M.~Eichner and V.~Ferrari.
\newblock Better appearance models for pictorial structures.
\newblock 2009.

\bibitem{eichner2010articulated}
M.~Eichner, M.~Marin-Jimenez, A.~Zisserman, and V.~Ferrari.
\newblock Articulated human pose estimation and search in (almost)
  unconstrained still images.
\newblock {\em ETH Zurich, D-ITET, BIWI, Technical Report No}, 272, 2010.

\bibitem{felzenszwalb2005pictorial}
P.~F. Felzenszwalb and D.~P. Huttenlocher.
\newblock Pictorial structures for object recognition.
\newblock {\em International Journal of Computer Vision}, 61(1):55--79, 2005.

\bibitem{Ferrari08}
V.~Ferrari, M.~Marin-Jimenez, and A.~Zisserman.
\newblock Progressive search space reduction for human pose estimation.
\newblock In {\em CVPR}, 2008.

\bibitem{fischler1973representation}
M.~A. Fischler and R.~A. Elschlager.
\newblock The representation and matching of pictorial structures.
\newblock {\em Computers, IEEE Transactions on}, 100(1):67--92, 1973.

\bibitem{girshick2014rcnn}
R.~Girshick, J.~Donahue, T.~Darrell, and J.~Malik.
\newblock Rich feature hierarchies for accurate object detection and semantic
  segmentation.
\newblock In {\em CVPR}, 2014.

\bibitem{gkioxariarticulated}
G.~Gkioxari, P.~Arbel{\'a}ez, L.~Bourdev, and J.~Malik.
\newblock Articulated pose estimation using discriminative armlet classifiers.
\newblock In {\em CVPR}, 2013.

\bibitem{ionescu2011latent}
C.~Ionescu, F.~Li, and C.~Sminchisescu.
\newblock Latent structured models for human pose estimation.
\newblock In {\em ICCV}, 2011.

\bibitem{Johnson10}
S.~Johnson and M.~Everingham.
\newblock Clustered pose and nonlinear appearance models for human pose
  estimation.
\newblock In {\em BMVC}, 2010.

\bibitem{Johnson11}
S.~Johnson and M.~Everingham.
\newblock Learning effective human pose estimation from inaccurate annotation.
\newblock In {\em CVPR}, 2011.

\bibitem{krizhevsky2012imagenet}
A.~Krizhevsky, I.~Sutskever, and G.~Hinton.
\newblock Imagenet classification with deep convolutional neural networks.
\newblock In {\em NIPS}, 2012.

\bibitem{mori2002estimating}
G.~Mori and J.~Malik.
\newblock Estimating human body configurations using shape context matching.
\newblock In {\em ECCV}, 2002.

\bibitem{nevatia1977description}
R.~Nevatia and T.~O. Binford.
\newblock Description and recognition of curved objects.
\newblock {\em Artificial Intelligence}, 8(1):77--98, 1977.

\bibitem{osadchy2007synergistic}
M.~Osadchy, Y.~LeCun, and M.~L. Miller.
\newblock Synergistic face detection and pose estimation with energy-based
  models.
\newblock {\em The Journal of Machine Learning Research}, 8:1197--1215, 2007.

\bibitem{pishchulin2013poselet}
L.~Pishchulin, M.~Andriluka, P.~Gehler, and B.~Schiele.
\newblock Poselet conditioned pictorial structures.
\newblock In {\em CVPR}, 2013.

\bibitem{ramanan2006learning}
D.~Ramanan.
\newblock Learning to parse images of articulated bodies.
\newblock In {\em NIPS}, 2006.

\bibitem{modec13}
B.~Sapp and B.~Taskar.
\newblock Modec: Multimodal decomposable models for human pose estimation.
\newblock In {\em CVPR}, 2013.

\bibitem{shakhnarovich2003fast}
G.~Shakhnarovich, P.~Viola, and T.~Darrell.
\newblock Fast pose estimation with parameter-sensitive hashing.
\newblock In {\em CVPR}, 2003.

\bibitem{sun2013deep}
Y.~Sun, X.~Wang, and X.~Tang.
\newblock Deep convolutional network cascade for facial point detection.
\newblock In {\em Computer Vision and Pattern Recognition (CVPR), 2013 IEEE
  Conference on}, pages 3476--3483. IEEE, 2013.

\bibitem{szegedy2013object}
C.~Szegedy, A.~Toshev, and D.~Erhan.
\newblock Object detection via deep neural networks.
\newblock In {\em NIPS 26}, 2013.

\bibitem{taylor2010pose}
G.~W. Taylor, R.~Fergus, G.~Williams, I.~Spiro, and C.~Bregler.
\newblock Pose-sensitive embedding by nonlinear nca regression.
\newblock In {\em NIPS}, 2010.

\bibitem{tian2012exploring}
Y.~Tian, C.~L. Zitnick, and S.~G. Narasimhan.
\newblock Exploring the spatial hierarchy of mixture models for human pose
  estimation.
\newblock In {\em ECCV}, 2012.

\bibitem{wang2013beyond}
F.~Wang and Y.~Li.
\newblock Beyond physical connections: Tree models in human pose estimation.
\newblock In {\em CVPR}, 2013.

\bibitem{yang2011articulated}
Y.~Yang and D.~Ramanan.
\newblock Articulated pose estimation with flexible mixtures-of-parts.
\newblock In {\em CVPR}, 2011.

\end{thebibliography}
}

\end{document}